\documentclass[lettersize,journal]{IEEEtran}
\usepackage{framed}
\usepackage{cite}
\usepackage{mathrsfs}
\usepackage{amsfonts}
\usepackage{amsmath}
\usepackage{makecell}
\usepackage{multirow}
\usepackage{graphics}
\usepackage{graphicx}
\usepackage{wrapfig}
\DeclareMathOperator*{\argmax}{arg\,max}

\usepackage{amssymb}
\usepackage{latexsym}
\usepackage{hyperref}
\usepackage{bbm} 

\usepackage{algorithm}
\usepackage{float}

\makeatletter

\newcommand{\Rmnum}[1]{\expandafter\@slowromancap\romannumeral #1@}
\makeatother
\begin{document}

    \title{Cardiac Evidence Backtracking for Eating Behavior Monitoring using Collocative Electrocardiogram Imagining}
    
    \author{
    Xu-Lu Zhang, Zhen-Qun Yang, Dong-Mei Jiang, Ga Liao, Qing Li,~\IEEEmembership{Fellow,~IEEE}, Ramesh Jain,~\IEEEmembership{Fellow,~IEEE}, Xiao-Yong Wei,~\IEEEmembership{Senior Member,~IEEE}
    \thanks{This work was supported by the National Natural Science Foundation of China  (61872256).}
    \thanks{
    Xiao-Yong Wei (e-mail: cswei@scu.edu.cn) and Xu-Lu Zhang are with the College of Computer Science, Sichuan University, Chengdu 610065, China}
    \thanks{Zhen-Qun Yang is with the Department of Biomedical Engineering, Chinese University of Hong Kong, Kowloon, Hong Kong (e-mail: jessicayang@cuhk.edu.hk).}
    \thanks{Xu-Lu Zhang and Dong-Mei Jiang are with the Center for Artificial Intelligence, Peng Cheng Lab, Shenzhen 518055, China.}
    \thanks{Dong-Mei Jiang is with the School of Computer Science, Northwestern Polytechnical University, Xi'an, China}
    \thanks{Ga Liao is with the State Key Laboratory of Oral Diseases, National Clinical Research Center for Oral Diseases, West China Hospital of Stomatology of Sichuan University, Chengdu 610041, China.}
    \thanks{Qing Li and Xiao-Yong Wei are with the Department of Computing, Hong Kong Polytechnic University, Kowloon, Hong Kong}
    \thanks{Ramesh Jain is with the Institute for Future Health, University of California, Irvine, CA, USA, and Donald Bren School of Information and Computer Sciences, University of California, Irvine, CA, USA.}
    \thanks{Zhen-Qun Yang is the co-first author.}
    \thanks{Xiao-Yong Wei is corresponding author.}
    }

	\maketitle	
	
	\begin{abstract}
		Eating monitoring has remained an open challenge in medical research for years due to the lack of non-invasive sensors for continuous monitoring and the reliable methods for automatic behavior detection.
		In this paper, we present a pilot study using the wearable 24-hour ECG for sensing and tailoring the sophisticated deep learning for ad-hoc and interpretable detection.
		This is accomplished using a collocative learning framework in which 1) we construct collocative tensors as pseudo-images from 1D ECG signals to improve the feasibility of 2D image-based deep models; 2) we formulate the cardiac logic of analyzing the ECG data in a comparative way as periodic attention regulators so as to guide the deep inference to collect evidence in a human comprehensible manner; and 3) we improve the interpretability of the framework by enabling the backtracking of evidence with a set of methods designed for Class Activation Mapping (CAM) decoding and decision tree/forest generation.
		The effectiveness of the proposed framework has been validated on the largest ECG dataset of eating behavior with superior performance over conventional models, and its capacity of cardiac evidence mining has also been verified through the consistency of the evidence it backtracked and that of the previous medical studies.

	\end{abstract}

	
	
	\section{Introduction}
	%
	%
	%
	%
	\IEEEPARstart{E}{ating} behavior logging is of great importance to eating pathology (e.g., obesity, binge eating, eating disorders) and essential to chronic disease management (e.g., diabetes, hypertension, and a wide range of other cardiovascular diseases). Users (patients) are usually required to record the time, duration, location, and food whenever they are taking any food. It is a tedious work which requires intensive cooperation from the users and is subject to recall bias. A lot of efforts towards automatic logging (monitoring) have been made recently (e.g., GPS for the location, food image recognition for dishes and ingredients \cite{jiang2018food}). However, the automatic eating monitoring (AEM), which detects the time and duration of eating in a non-invasive way, has not been addressed well. AEM is considered as the basis for automatic logging, because it happens frequently in the clinical experiments that users forget the time and duration. Furthermore, the other techniques like GPS recording and food recognition, which request on-the-fly actions from users (to record immediately when the eating happens), cannot be applied without AEM.
	
	\begin{figure*}[t]
		\centerline{\includegraphics[width=0.58\textwidth]{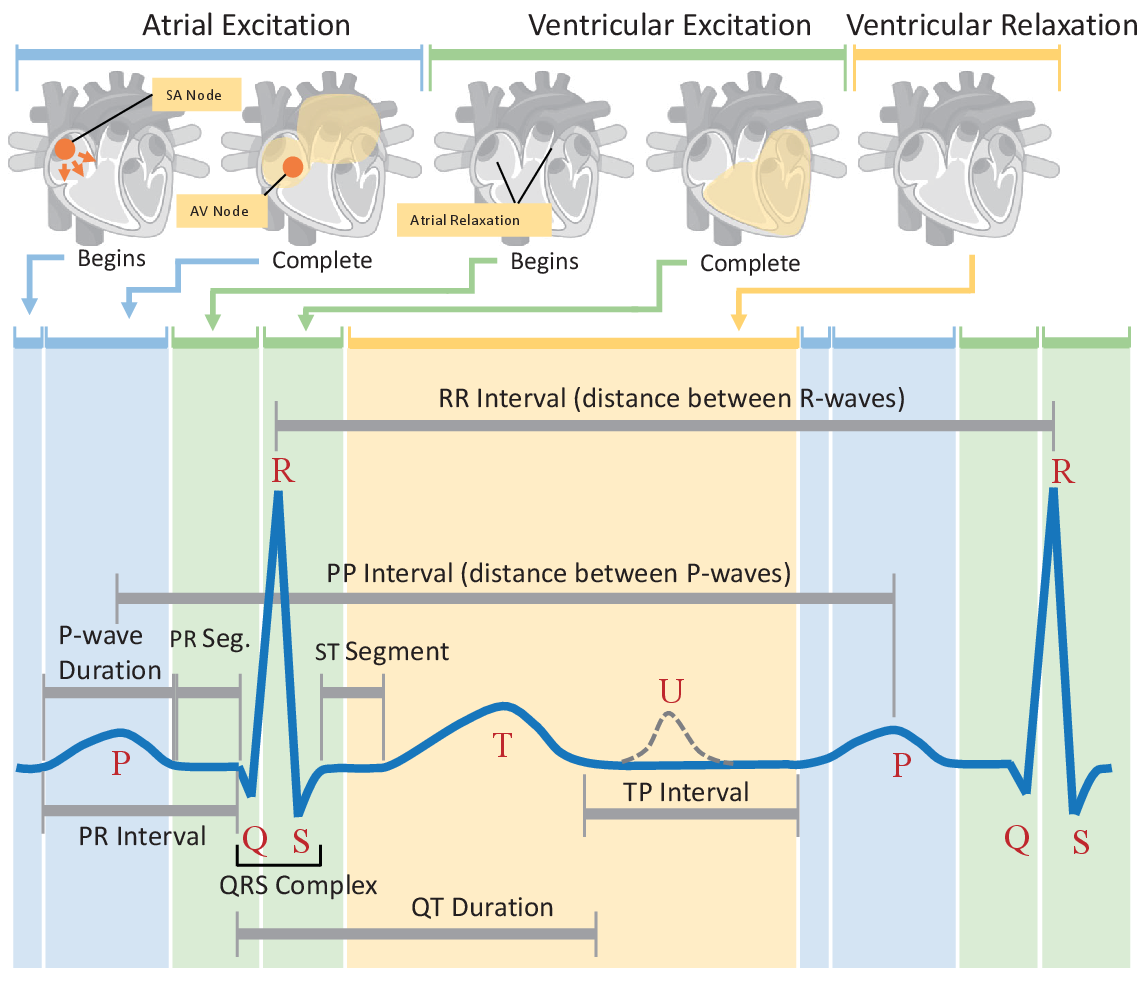}}
		\caption{Illustration of ECG signals and the typical intervals for cardiac analysis. An ECG period includes 6 waves that are related to different parts of the heart where the electrical impulse initiation, conduction, and depolarization have been performed (e.g., P wave indicates the impulse conduction from the sino-atrial node to the atrio-ventricular node). However, in cardiology, the interpretation of ECG has been built more on the comparative features like the morphology and intervals of these waves (e.g., PP, QRS) than the values of a single wave or time point, because a heart event is usually a dynamic process that includes a set of consecutive movements of several heart parts. This makes the classical features in signal processing (e.g., zero-crossing rates) play a less effective role than the comparative features. 
			In other words, the cardiology way of interpretation focuses on the inter-wave binary relation rather than the unary relation (of single wave).
			%
		}\label{fig:ECGwave}
	\end{figure*}
	
	Studies show that the eating activity is closely linked to heart movement as food digestion will affect how the heart pumps blood \cite{hnatkova2014qtc,scott2002carbohydrate,taubel2012shortening}. However, these studies are still far from creating practical methods for automatic AEM, because they have been limited in the scope of qualitative medical research.
	The 24-hour Electrocardiogram (ECG) monitoring is an emerging technique to track the continuous heart activity for cardiovascular health problems, and thus theoretically can be used for AEM.
	Nevertheless, our initial study shows it only gives a moderate performance by using the features found in previous qualitative studies (e.g., heart rates or heart rate variability) \cite{hnatkova2014qtc}.
	
	To build a better detector based on the 24-hour ECG, one may naturally think of it as a signal processing problem which can be solved using the advanced deep learning methods. Despite its success in other signal processing applications, this handy solution does not make much sense in a cardiology point of view, because the cardiology analysis is based more on the \textit{comparative} features between cardiac waves (as shown in Fig.~\ref{fig:ECGwave}) rather than the statistical features used in signal processing. Furthermore, in cardiac analysis, features or models need to be \textit{interpretable}, because either the waves themselves or the inter-wave relation (e.g., intervals) have their cardiac seedbed, and it is more important for a model to locate the evidence than to make the prediction. This makes the nontransparent deep models less feasible to this problem (even the high accuracy of prediction might be obtained). Moreover, there is an even more fundamental issue which lies in the 1D and \textit{periodic} nature of the ECG data, on which the majority of the sophisticated deep models (e.g., VGG \cite{simonyan2014very}, ResNet \cite{he2016deep}, Inception-v3 \cite{szegedy2016rethinking}) cannot play to their strength, because these models are built on 2D image-based data and the periodic characteristics have not been considered in the design.
	
\begin{figure*}[t]
		\centering
		\includegraphics[width=0.97\textwidth]{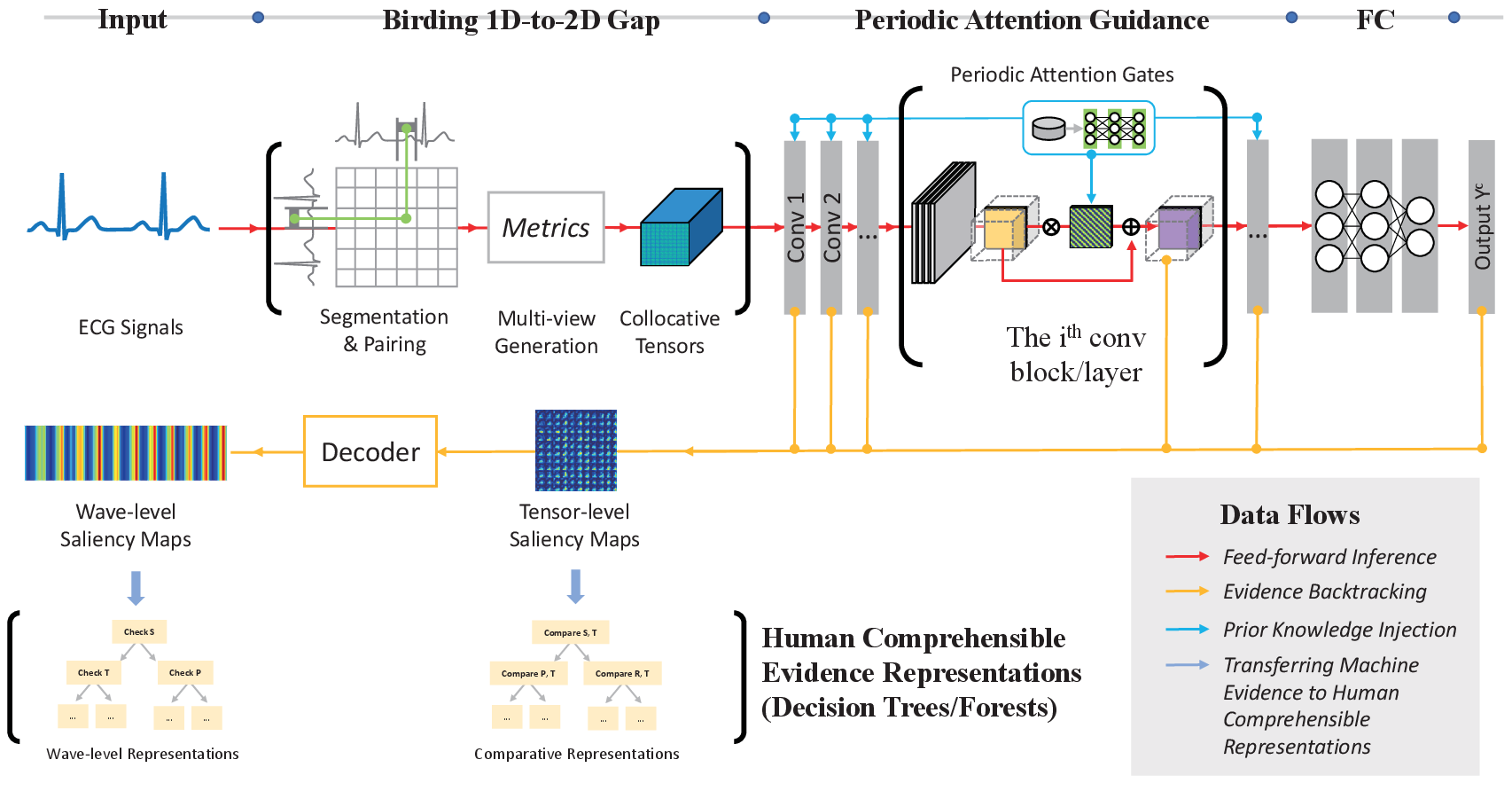}
		\caption{The collocative framework for modeling the comparative relations and cardiac evidence backtracking in Electrocardiograms. It consists of three parts: 1) the transformation of 1D ECG signals into 2D views to be adapted to the conventional deep models; 2) the periodic attention coach which guides the deep inference with human logic of comparative analysis; 3) the evidence backtracking based on saliency maps and the generation of human comprehensible representations of the evidence and inference logic.}\label{fig:framework}
\end{figure*}

	In this paper, we present a pilot study of using 24-hour ECG for non-invasive and continuous eating behavior monitoring. A deep learning framework has been proposed to guide the learning towards human comprehensible inference and thus enable the cardiac evidence backtracking. 
	The framework has addressed the aforementioned issues (i.e., comparative analysis, interpretability, periodic nature, and 1D-to-2D gap). More specifically, we have built a new collocative learning \cite{wei2021deep} scheme to transfer the 1D ECG signals into 2D pseudo-images which encapsulates the comparative features among ECG segments, so that we can bring the sophisticated deep models into full play. In addition, a coaching attention mechanism has been designed to adapt these models to the periodic nature of ECG signals. More importantly, we will show that within the framework, we are able to backtrack the cardiac evidence of the predictions in the form of deep learning favored saliency maps and human comprehensible representations.
	%
	%
	The framework is shown in Fig.~\ref{fig:framework}. It has advantages over existing methods as follows
	\begin{itemize}
		\item \textbf{Comparative Relation}: We transform the 1D ECG signals into collocative tensors (based on 2D views) in which the comparative relations have been encapsulated with the inter-segment correlation/distance. The tensors are pseudo-images which are compatible to the majority of deep models. It gives us the opportunity of analyzing the ECG data in a (sophisticated) image liked way. Moreover, we will show in Section~\ref{sec:multi-view_construct} that the multi-view setting is able to create a more comprehensive representation for deep learning.
		\item \textbf{Period Sensing}: With the multi-view tensors, we will see in Section~\ref{sec:periodic_attention} that the morphological patterns of waves will be converted into ``backslashes'' in the pseudo-images, on the basis of which we have designed a periodic attention template that is able to adapt its period to the distance between or length of the ``backslashes''. This has transformed the problem of learning the period of signals into a distance/length sensing problem which is easier to be addressed with image-based deep models.
		\item \textbf{Interpretable Evidence Backtracking}: With the periodic attention template, we can guide the deep models to select only ``backslashes'' that repeat with a certain period so as to imitate the human logic of evidence building. This will be reflected in both the tensor- and wave-level saliency maps.
		Consequently, by decoding the saliency maps, we are able to backtrack the evidence (i.e., the comparative features of wave pairs) and finally represent the evidence and inference logic in a human comprehensible way (i.e., decision trees/forests), which can help the technicians to identify cardiac seedbed of the eating behavior.
	\end{itemize}
	The contribution of this paper includes: we have conducted a pilot study of ECG-based eating behavior logging/monitoring, and have brought image-based deep learning methods into this area for comparative relation mining and interpretable evidence backtracking.
	In addition, we will also release the first and largest ECG dataset for eating behavior analysis to promote the research in this area.

	\section{Related Work}
	\subsection{Wearable Eating Behavior Monitoring}
	Eating is essential for the health of human beings. The monitoring of eating is thus of great importance to either health management of the general population or eating pathology for patients. Through its long history of thousands of years, the monitoring has been mainly based on self-reporting, which is subjective and with recall bias \cite{basiotis1987number}. This brings forward automatic methods especially those using wearable devices. 
	A wide range of efforts has been made in the last two decades \cite{amft2010wearable,dong2012new,farooq2014novel,farooq2015comparative,liu2012intelligent,passler2012food,sazonov2008non,sun2014ebutton,vu2017wearable}. Various sensors have been applied, such as acoustic sensors using microphones \cite{amft2010wearable}, visual sensors using cameras \cite{sun2014ebutton}, wrist or arm motion sensor using accelerometers \cite{dong2012new}, physiological sensors using EMG \cite{farooq2014novel}, piezoelectric sensors \cite{farooq2015comparative}, and multi-sensor by combining these sensors \cite{liu2012intelligent}.
	
	The choice of sensors is in fact a choice of balance between the comfort and reliability. For example, microphones have been widely used for monitoring by recognizing the sound of chewing and swallowing. The outer-ear \cite{passler2012food} or neck \cite{sazonov2008non} microphones might be comfort to wear, but less reliable due to the disturbance of the environmental noise. The inner-ear microphones are reliable but less user friendly.
	In addition, in most applications (especially when users are patients), more important is that whether the sensors can provide pathological evidence for more in-depth analysis of the users' condition. The EMG or piezoelectric sensors are good in this regard, because they are connected to the inner processes (e.g, muscle movement and tension) of the body. However, these sensors are less acceptable to users as they have to be installed around the neck.
	
	In this paper, we study the possibility of using the 24-hour ECG sensors. As a non-invasive sensor, it is comparatively comfort, and most importantly, it has been widely used in collecting both the
	physiological and psychological evidence \cite{serhani2020ecg}, because it is associated to the heart which is one of the most important organs of our body and determines both the physical and mental health.

	\subsection{Deep Learning for ECG Analysis}
	In terms of learning methods employed for eating monitoring, although neural network-based models have also been explored by a few researchers \cite{farooq2013comparative}, conventional models, such as HMM \cite{bi2015autodietary}, SVM \cite{lopez2010detection}, random forests \cite{fontana2013estimation}, are still dominating in this area, because it is straightforward to consider the monitoring as a 1D signal processing (detection) problem. Early work in \cite{farooq2013comparative} has used a 3-layer neural network 
	which outperforms SVM, but it remains as a conventional shallow network which has not leveraged the power of deep learning.
	
	Beyond the task of eating behavior monitoring, deep learning has already been adopted in a number of studies for ECG analysis \cite{al2018convolutional,andreotti2017comparing,fan2018multiscaled,hannun2019cardiologist,he2018automatic,naz2021ecg,sarkar2020self,ullah2020classification}. A majority of them are for arrhythmia classification \cite{andreotti2017comparing,hannun2019cardiologist} together with a few for emotion recognition \cite{sarkar2020self} or atrial fibrillation detection \cite{fan2018multiscaled}. Main-stream methods are either to feed the 1D ECG signals to the 1D CNNs (ResNet or VGG) directly \cite{andreotti2017comparing,hannun2019cardiologist,fan2018multiscaled} or to adopt 2D CNNs after transforming the 1D inputs into 2D spectrograms or simply reshaping the signals into 2D \cite{al2018convolutional,he2018automatic,naz2021ecg,ullah2020classification}. 
	None of these methods has modeled the periodic nature of the signal. Furthermore, 
	none of these methods has modeled the high order relation (inter-wave comparison) specifically, and thus is limited to provide explicit evidence for cardiac analysis. 
	In this paper, we will address this problem with collocative learning which is capable of period sensing and evidence backtracking. This has been done through a periodic attention coaching mechanism and a CAM-based decoding process for backtracking and human comprehensible representation generation.
	
	\subsection{Attention and CAM Methods}
	Attention mechanisms have been employed in deep learning for weighting the feature maps so that the highly important regions can stand out from the feed-forward inference \cite{hu2018squeeze,wang2018non,woo2018cbam}. The attention usually appears as a weighting mask which has been learned through back-propagation adaptively, meaning the mask formation is data-driven, network dependent, and without human guidance. The coached attention gates (Section~\ref{sec:cag_intro}) we proposed is different from standard attention mechanisms in a way that we define a parameterized template mask with embedded periodic patterns and allow the back-propagation to learn the parameters only rather than the mask directly. It is a combination of prior knowledge injection and adaptive learning.
	
	CAM-based methods are often used to generate saliency maps which indicate which parts of the inputs have played a more significant role in the inference than others \cite{chattopadhay2018grad,selvaraju2017grad,zhou2016learning}. However, the saliency maps are usually with an arbitrary shape for providing indicative rather than explicit evidence (e.g., exact localization of waves or wave pairs). In this paper, we have built a decoding scheme (see Section~\ref{sec:evidence_backtracking}) on top of the Grad-CAM \cite{selvaraju2017grad}, which is able to convert the indicative evidence into explicit evidence for generating human comprehensive representations.
	
	In terms of collocative learning, this paper is also related to \cite{wei2021deep} in which the prototype framework of collocative learning has been proposed. In this paper, we have followed the thread but have implemented the learning in a different way (see Section~\ref{sec:relation_model}). Most importantly, this framework in this paper has been designed specifically for the ECG-based eating behavior monitoring which is a completely different task from the Immunofixation Electrophoresis analysis in \cite{wei2021deep}.
	

	\section{Comparative Relation Modeling and Cardiac Evidence Backtracking}
	\label{sec:relation_model}
	In this section, we will depict the proposed collocative learning framework consisting of three parts: 1) the transformation of the 1D ECG signals into 2D views to adapt to image-based deep models, 2) the knowledge injection through attention regulation to guide the deep networks to conduct the inference by integrating both the periodic nature of ECG signals and human logic of comparative analysis, and 3) the evidence backtracking and the generation of human comprehensible representations.
	
	\subsection{Multi-view ECG Tensor Construction}
	\label{sec:multi-view_construct}
	\subsubsection{Inter-Segment Relation Encoding}
	It is straightforward to convert a 1D signal $S$ into an image by cutting a signal into $n$ segments and then encapsulating the inter-segment correlations/distances into a $n\times n$ matrix $\boldsymbol{R}$. However, it will lose the original 1D features despite the inter-segment relation has been encoded. To address this issue, we can encode the original features into the diagonal of the matrix $\boldsymbol{R}$. It is a near-lossless replacement as the diagonal elements are all ones (correlation) or zeros (distance). It can be formulated as
	\begin{equation}\label{eq:relation_matrix}
		\begin{split}
			\boldsymbol{R}=&\underbrace{\boldsymbol{R}\odot(1-\boldsymbol{E})}_{\text{diagonal mask-out}}\hspace{0.5cm}+\hspace{0.1cm}\underbrace{(\boldsymbol{\vec{1}}\boldsymbol{\vec{x}}^\top)\odot\boldsymbol{E}}_{\text{segment feature encoding}}\\
			\boldsymbol{\vec{x}}=&f(S), S=\{s_i|1\leqslant i\leqslant n,i\in \mathbb{Z}^+\}
		\end{split}
	\end{equation}
	where the $\odot$ is the Hadamard product, $\boldsymbol{E}$ is a $n \times n$ identity matrix, $\boldsymbol{\vec{1}}$ is a $n \times 1$ all-ones vector, and $\top$ denotes transpose operator. The $\boldsymbol{\vec{x}}$ is a $n \times 1$ feature vector where the $i^{th}$ element is the original feature value of the $i^{th}$ segment of the ECG signal $S$ which can be extracted by any convectional signal processing function $f(\cdot)$. Eq.~(\ref{eq:relation_matrix}) equals to a process that ``masks out'' the diagonal of $\boldsymbol{R}$ and then replaces with the original segment features.
	
	\subsubsection{Composing Relation Matrices into A Tensor}
	The resulting $\boldsymbol{R}$ is indeed a function of the metric used to calculate the correlations/distances (denote as $\phi(s_i,s_j)$ hereafter) and the feature extraction function $f(s_i)$. It is more representative than the original data because of the binary and comparative relation (of two segments) extracted by $\phi(s_i,s_j)$ and the unary relation (of a single segment) extracted by  $f(s_i)$.
	By varying $\phi(\cdot)$ and $f(\cdot)$, we can generate a set of $\boldsymbol{R}$s that represent our understanding about the original ECG data from different point of views. The multi-view tensor $\boldsymbol{\mathcal {T}}$ can then be defined as an order-3 tensor composed by the concatenation of these $\boldsymbol{R}$s as
	\begin{equation}\label{eq:concat_single_view}
		\boldsymbol{\mathcal {T}}=\{\boldsymbol{R}_{f_1,\phi_1},\boldsymbol{R}_{f_2,\phi_2},\dots,\boldsymbol{R}_{f_m,\phi_m}\}=\bigcup_{i=1}^{m}\boldsymbol{R}_{f_i,\phi_i}
	\end{equation}
	where $m$ denotes the number of views. The $f_i$ can be selected from a wide range of sophisticated signal processing functions, and the $\phi_i$ can be implemented by metrics such as cosine similarity, Euclidean distance, Manhattan distance, and Mahalanobis distance. 
	
	\subsubsection{Rationale Behind The Tensor Design}
	$\boldsymbol{\mathcal {T}}$ is an image-liked input (or pseudo-image) which fits the input format of most of the DNN models. It is with dimensions $n\times n\times m$ in which the first two dimensions ($n\times n$) are imitating the spatial layout of an image and the relation among elements will be ``preserved'' so that we can decode from the CAM-based saliency maps for figuring out which pairs of segments have played a more important role than others. By contrast, the last dimension ($m$) is imitating channels which will be fused during the feed-forwarding process.
	We will see in Section~\ref{sec:experi_tensor_construct} that the fusion can be considered as a multi-view learning that has the advantage of combining the complementary information from various views through an unified constraint of attention.
	
	\subsection{Periodic Attention Regulation}
	\label{sec:periodic_attention}
	With the multi-view tensor as the input to deep networks, we can build an attention regulator to guide the inference focus more on the comparative features with the periodic nature of ECG signal taken into consideration.
	
	\subsubsection{Periodic and Comparative Features Encoded in \texorpdfstring{$\boldsymbol{\mathcal {T}}$}{}}
\begin{figure}[t]
		
		\centerline{\includegraphics[width=0.9\columnwidth]{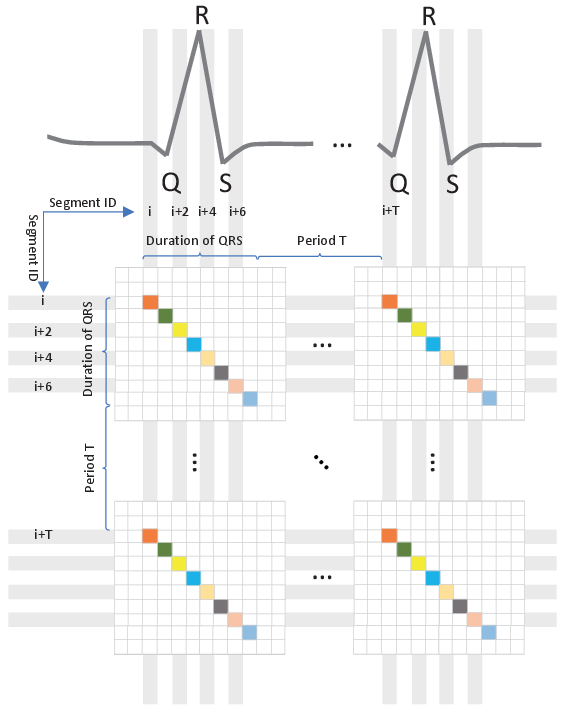}}
		\caption{The comparative relations among segments have already been encoded as ``backslashes'' into the collocative tensors. In the figure, several backslashes for the QRS complex are shown. It can be used to model the comparative relations of Q, R, and S waves. The duration has also been encoded as the length of backslashes. The period is encoded into the distances among backslashes.}\label{fig:Tensor_construct}
	\end{figure}

	The first challenge is that there is no existing image-based attention mechanism which has specifically addressed the periodic nature of the original signal. By reviewing our input tensor $\boldsymbol{\mathcal {T}}$, we can see that the signal period (denote as $T$ hereafter) has already been encoded, in a way that the correlation/distance will also repeat while the signal repeats, or in a formulated way as
	\begin{equation}\nonumber
		\begin{aligned}
			&\mathbb{E}[f(s_{i+kT})-f(s_i)]=0,0\leqslant k\in \mathbb{Z}\\
			&\hspace{3cm}\Updownarrow\\
			&\mathbb{E}[\phi(s_{i+kT},s_{j+lT})-\phi(s_{i},s_{j})]=0,0\leqslant k,l\in \mathbb{Z}
		\end{aligned}
	\end{equation}
	where the $\mathbb{E}[\cdot]$ denotes the expectation. 
	Taking this back to our input tensor $\boldsymbol{\mathcal {T}}$, we will see in Fig.~\ref{fig:Tensor_construct} that if a comparative pattern of the 1D signal is repeating consistently, it will form in $\boldsymbol{\mathcal {T}}$ multiple ``backslashes'' with distances among them indicating the periods. 
	
	Let us review how a human technician detects such patterns for a more intuitive explanation through an example.
	Among the 6 ECG waves illustrated in Fig.~\ref{fig:ECGwave}, the QRS complex consisting 3 waves of Q, R, and S is one of the most frequently used ``makers'' for the cardiovascular examination, with which a technician can identify from its morphology whether the depolarization (activation) of the ventricles functions properly. More specifically, a QRS usually has a unique appearance with a sharp peak at the R and two short valleys at the Q and S. As shown in Fig.~\ref{fig:Tensor_construct}, this appearance is indeed a trend change of consecutive segments which can be captured by their correlation/distance stored in the input tensor $\boldsymbol{\mathcal {T}}$ where 
	its values from the onset to the end of a QRS form a ``backslash''. The backslash repeats itself following the period $T$ resulting in parallel backslashes. It is a distinguishable pattern that can be easily learned by most of the CNN models designed for image-liked inputs.
	Furthermore, the duration of a QRS in fact indicates how strong and fast the ventricles contract to pump blood to other parts of the body. Generally speaking, a normal QRS is short (or narrow) while a QRS within a ventricular dysrhythmia is wide (or broad). This will be reflected by the length of the QRS backslashes in $\boldsymbol{\mathcal {T}}$. 
	%
	Moreover, we can find from Fig.~\ref{fig:Tensor_construct} that the period $T$ can be measured straightforwardly from the distances of the ``backslashes''.
	However, CNN models are usually not specifically designed to learn such distances and their variations. We can address this issue by adding the attention regulation on the CNN inference.
	
	
	\subsubsection{Attention Regulation with Coached Attention Gates}
	\label{sec:cag_intro}
	We have learned that there are two characteristics of the tensor encoding that we need to consider in our attention mechanism design, namely the back-slashed patterns and the distances of backslashes. Let us rewrite the challenges in a more specific way as
	\begin{itemize}
		\item Periodic backslash modeling: Although the backslashes themselves are with distinguishable appearance and easy to be modeled in traditional CNNs, we need the model to focus on backslashes that repeat regularly with a certain period. This has not been addressed by traditional CNNs.
		\item Distance sensing: In addition, we need the model to be able to ``sense'' the distance and its variations between backslashes so that the intervals can be modeled.
		\item Period variation: Furthermore, beside the PP/RR intervals, technicians are actually referring to a lot of other intervals for ECG interpretation (e.g., PR, QT, QS, ST). These intervals are often with their own inherent periods which vary from the overall $T$ (usually measured through the heart beat rhythm) with little variances in different degrees.
	\end{itemize}
	
	We propose to address these issues with the Coached Attention Gates (CAGs). The idea is to use a diagonalized attention mask to coach the networks to only search evidence at the diagonal direction (for backslash modeling),  then an attention gate (i.e., a CAG unite) can be formed by repeating the mask with an adaptive period learned from the back-propagation (for distance sensing), and finally we can apply the principal to different layers/blocks of the networks and generate CAG unites to adapt to the period variation at different degrees (for period variation).
	
	{\textbf{CAG Intuition:}}
	Let us denote a CAG unite as $\boldsymbol{\Omega}$ that is a matrix with the dimensions ($m\times m$) which are the same as those of the spatial dimensions of a target feature map $\boldsymbol{\mathcal{F}}$ (of shape $m\times m\times Z$, $Z$ denotes the number of channels). The attention regulation can be written as   
	\begin{equation}\label{eq:regulation}
		\boldsymbol{\mathcal{F}}^*_k=\boldsymbol{\mathcal{F}}_k\odot \boldsymbol{\Omega} + \boldsymbol{\mathcal{F}}_k, 1\leqslant k\leqslant Z
	\end{equation}
	where the $k$ is an iterator to traversing channels of $\boldsymbol{\mathcal{F}}$.

	\begin{figure}[t]
		
		\centerline{\includegraphics[width=0.97\columnwidth]{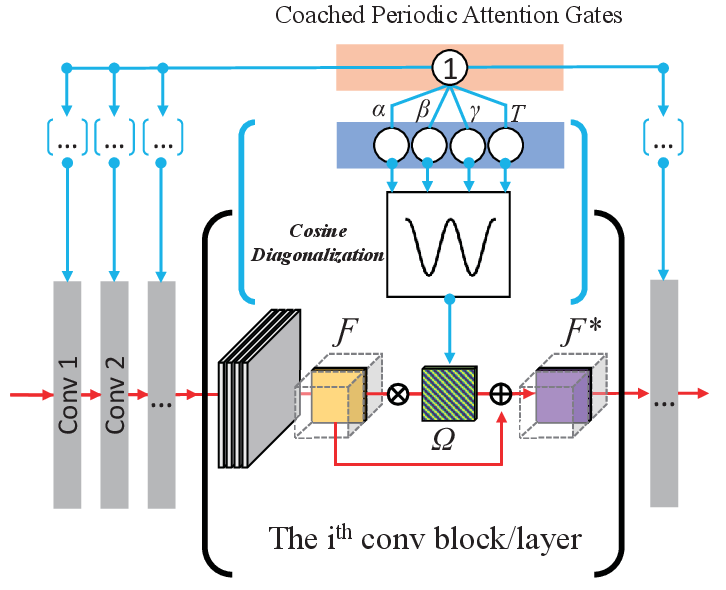}}
		\caption{The coached periodic attention gates (CAGs). The network of CAGs consists of three layers. The first layer is a single node fixed value 1. The second layer is with $4$ parameter nodes of $(\alpha,\beta,\gamma,T)$. The third layer is a parameterized (by the parameters of the second layer) and diagonalized cosine attention mask $\boldsymbol{\Omega}$ which will guide the learning to focus on the diagonal ``backslash'' patterns and sense the period of the signals at the same time.}\label{fig:cag_module}
	\end{figure}

	The CAG unite can be implemented with a trigonometric function (e.g., \textit{sine}, \textit{cosine}) along the diagonal, because it is easy to parameterize the function to control its amplitude and period. An illustration of CAG generated using a parameterized \textit{cosine} is shown as the $\boldsymbol{\Omega}$ in Fig.~\ref{fig:cag_module}, from which the coaching process becomes intuitive. 
	Firstly, the diagonalization of the attention ``strips'' has confined the way of the networks to collect evidence only from backslash patterns. Secondly, the repetition of the strips has regulated the inter-backslash relation to be periodic. In other words, the more consistent the repetition periods of the strips and backslashes are, the larger ``resonance'' (i.e., agreement) between the CAG and the feature map reaches, which thus leads to the more robust evidence for inference. Furthermore, the agreement is also an indication of better consistency between the human logic (formulated through the CAG) and the deep network inference. Otherwise, the deep network may collect evidence from patterns of arbitrary shapes if there is no such CAG regulation. Nonetheless, in spite of the rationale of the intuition, we should see that the period $T$ for the \textit{cosine} is not easy to be defined because of the period variation we have introduced. We will leverage the power of deep learning to ``sense'' it and also make it adaptive to the variation. Let us formulate CAG to see how this has been implemented.
	
	{\textbf{CAG Formulation:}} The values of a CAG unite $\Omega$ can be obtained from a function $\omega(i,j|\alpha,\beta,\gamma,T)\in[0,1]: \mathbb{Z}^+\times \mathbb{Z}^+\to \mathbb{R}$ that assigns each location $(i,j)$ an attention value $\boldsymbol{\Omega}_{ij}=\omega(i,j)$ based on a set of learned parameters $(\alpha,\beta,\gamma,T)$ controlling the shape (and offsets) of the \textit{cosine} wave. Putting all together, it is given as
	\begin{equation}
		\begin{split}
			\boldsymbol{\Omega}_{i,j}=&\omega(i,j|\alpha,\beta,\gamma,T)\\
			=&\underbrace{\alpha}_{\text{scalar}}\cdot\boldsymbol{\cos}\left(2\pi T\underbrace{\frac{|i-j|}{m}}_{\text{normalizer}}+\underbrace{\beta}_{\text{offset}}\right)+\underbrace{\gamma}_{\text{adjuster}}   
		\end{split}
	\end{equation}
	where $\alpha$ is a scalar on the cosine to control the amplitude of the attention, $\frac{|i-j|}{m}$ is a normalizer to convert the spatial locations into the range of the input field of the \textit{cosine}, $\beta$ controls the offset of the cosine, and $\gamma$ helps the attention values to fit into a proper range so that it can be used in Eq.~(\ref{eq:regulation}). Note that from an attention mask point of view, the $\alpha$ determines the weight range between these highly trusted evidence (backslashes) and the less trusted ones, while $\beta$ determines how the mask shifts along the anti-diagonal so that the signal input can start at any given point of a cardiac cycle. This setting has released the requirements of detecting the onset of P waves and segmenting the signal into standard cardiac waves, and thus makes the process more feasible, because cardiac wave/peak detection is an open question and ongoing study by itself \cite{laitala2020robust,vijayarangan2020rpnet}. 
	
	{\textbf{CAG Learning:}} The parameter set $(\alpha,\beta,\gamma,T)$ is not easy to learn in a conventional setting of signal process or cardiac analysis. Fortunately, we can leverage the power of deep learning to figure them out. The CAGs will be organized as an attention network as shown in Fig.~\ref{fig:cag_module}. The first layer is with only one node of value $1$ as the common root of the network. The second layer is a parameter layer consisting of $L$ CAG blocks where $L$ equals to the number of blocks/layers of the host CNN model. Each block is with $4$ nodes for $\alpha$, $\beta$, $\gamma$, and $T$ respectively. The third layer is a gate generator layer for creating $H$ CAG masks ($\Omega$s) with parameters of the second layer. The resulting CAGs will be planted into each convolutional layer/block of the host CNN model to implement the attention regulation in Eq.~(\ref{eq:regulation}). In each CAG block, mask nodes in the third layer are fully connected to the parameter nodes in the second layer, so that the parameters ($\alpha$, $\beta$, $\gamma$, and $T$) can be then learned during the back-prorogation. However, we have to modify the gradient updating functions for the second layer since the \textit{cosine} used for connecting nodes between layers is a non-standard DNN schema. With the chain rules, we define the gradients of the $(i,j)$ node of the $\boldsymbol{\Omega}$ regarding current prediction $Y^c$ for the class $c$ as
	
	\begin{flalign}
		\begin{split}
			\frac{\partial Y^c}{\partial \alpha}&=\sum_{i,j}\frac{\partial Y^c}{\partial \Omega_{ij}}\frac{\partial \Omega_{ij}}{\partial \alpha}\\
			&=\sum_{i,j}\frac{\partial Y^c}{\partial \Omega_{ij}}\frac{\partial \alpha\cos\left(2\pi T\frac{|i-j|}{m}+\beta\right)+\gamma}{\partial \alpha}\hspace{2cm}\\
			&=\sum_{i,j}\frac{\partial Y^c}{\partial \Omega_{ij}}\cos\left(2\pi T\frac{|i-j|}{m}+\beta\right),\\
		\end{split}
	\end{flalign}
	
	\begin{flalign}
		\begin{split}
			\frac{\partial Y^c}{\partial T}&=\sum_{i,j}\frac{\partial Y^c}{\partial \Omega_{ij}}\frac{\partial \Omega_{ij}}{\partial T}\\
			&=\sum_{i,j}\frac{\partial Y^c}{\partial \Omega_{ij}}\frac{\partial \alpha\cos\left(2\pi T\frac{|i-j|}{m}+\beta\right)+\gamma}{\partial T}\hspace{2cm}\\
			&=\sum_{i,j}\frac{\partial Y^c}{\partial \Omega_{ij}}\left[-\alpha\sin\left(2\pi T\frac{|i-j|}{m}+\beta\right)\frac{2\pi|i-j|}{m}\right]\\	
			&=-\sum_{i,j}\frac{\partial Y^c}{\partial \Omega_{ij}}\alpha\sin\left(2\pi T\frac{|i-j|}{m}+\beta\right)\frac{2\pi|i-j|}{m},\\
		\end{split}
	\end{flalign}
	
	\begin{flalign}
		\begin{split}
			\frac{\partial Y^c}{\partial \beta}&=\sum_{i,j}\frac{\partial Y^c}{\partial \Omega_{ij}}\frac{\partial \Omega_{ij}}{\partial \beta}\\
			&=\sum_{i,j}\frac{\partial Y^c}{\partial \Omega_{ij}}\frac{\partial \alpha\cos\left(2\pi T\frac{|i-j|}{m}+\beta\right)+\gamma}{\partial \beta}\hspace{2cm}\\
			&=\sum_{i,j}\frac{\partial Y^c}{\partial \Omega_{ij}}\left[-\alpha\sin\left(2\pi T\frac{|i-j|}{m}+\beta\right)\right]\\
			&=-\sum_{i,j}\frac{\partial Y^c}{\partial \Omega_{ij}}\alpha\sin\left(2\pi T\frac{|i-j|}{m}+\beta\right)\\
		\end{split}
	\end{flalign}
	
	where $\frac{\partial Y^c}{\partial \Omega_{ij}}$ and $\frac{\partial Y^c}{\partial \Omega_{ij}}$ can be learned from the standard back-propagation.
	Note that the period parameter $T$ learned in each block varies from each other. It generates the maximum ``resonance'' to support the inference when the $T$s at the early layers approach the cardiac period of the original ECG signal. For the other $T$s learned thereafter, they will adapt to the scale change caused by the convolution. This will help the inference to ``sense'' the period at different scales so that the period variation issue can be addressed.
	
	\subsection{Evidence Backtracking and Generation of Human Comprehensible Representations}
	\label{sec:evidence_backtracking}
	In CNN-based deep networks, we can generate the saliency maps using CAM-based visualization methods and investigate which evidence has been used to support the inference, because the saliency maps are considered as an indication of the activation of the feature maps. However, in conventional saliency maps, it can only tell which parts are more important than others. It is hard to find how two parts are used jointly to provide comparative evidence. In collocative learning, we can extract such evidence because the comparative relations among segments has already been ``wrapped'' into the input tensors. In this section, we will ``decode'' the collocative saliency maps and use it to generate human comprehensible evidence representations to assist the cardiac analysis.
	
	\subsubsection{Collocative Saliency Map Generation}
	We can simply generate a saliency map $\boldsymbol{\mathcal{S}}^{\boldsymbol{\mathcal{F}}^*}_l, 1\leqslant l \leqslant L$ using Grad-CAM \cite{selvaraju2017grad} for each of the $L$ layers/blocks of the host CNN model based on the feature map $\boldsymbol{\mathcal{F}}^*$, and then integrate these maps into the final map using average pooling. It is written as
	
	\begin{equation}
		\boldsymbol{\mathcal {S}}=\frac{1}{L}\bigoplus_{l=1}^L\mathcal {S}^{\boldsymbol{\mathcal{F}}^*}_{\{l\}}.
	\end{equation}
	The resulting $\boldsymbol{\mathcal {S}}$ can be resized to match the spatial dimensions of the collocative tensor $\boldsymbol{\mathcal {T}}$, so that the significance of the comparative feature of any segment pair $(s_i, s_j)$ in the inference can be evaluated quantitatively. 
	
	\subsubsection{Decoding The Maps Into Wave Scale}
	The segments that represent a machine perceptible way of data reading are with a much finer scale than that of the cardiac waves used by human technicians. We need to decode the segment-based saliency map $\boldsymbol{\mathcal {S}}$ into wave-based rating to help the human interpretation of the results. With the collocative configuration of $\boldsymbol{\mathcal {T}}$, the map can be decoded into either the rating of unary waves or that of the binary wave-to-wave comparison.
	
	Assume we have a set of waves $W=\{w_p|1\leqslant p\leqslant |W|,p\in \mathbb{Z}^+\}$ and another set of segments $S=\{s_i|1\leqslant i\leqslant |S|,i\in \mathbb{Z}^+\}$, we can decode the segment-based saliency map $\boldsymbol{\mathcal {S}}$ into a rating of unary waves by a voting schema in which the weight $\boldsymbol{\mathcal {S}}_{ij}$ (the significance of comparing $s_i$ and $s_j$) will be assigned to the waves $w_p$ and $w_q$ respectively if $s_i\in w_p$ and $s_j\in w_q$.
	This can be implemented as
	\begin{equation}\label{eq:saliency_vec}
		\begin{split}
			\boldsymbol{\vec{s}}_W=&\frac{1}{|S|}
			\underbrace{\vphantom{\frac{1}{2\|\vec{\mathbf{1}}\| }}\boldsymbol{\mathcal{M}}^\top}_{\text{membership}}
			\underbrace{\frac{1}{2\|\vec{\mathbf{1}}\|}\left(\boldsymbol{\mathcal {S}}\vec{\mathbf{1}}+\boldsymbol{\mathcal {S}}^\top\vec{\mathbf{1}}\right)}_{\text{segment-based voting}},\\
			\boldsymbol{\mathcal{M}}_{ip}=&\left\{
			\begin{array}{cc}
				1,& s_i\in w_p\\
				0,& s_i\notin w_p
			\end{array}
			\right.,
		\end{split}
	\end{equation}
	where the $\boldsymbol{\mathcal{M}}_{ip}$ is a $|S|\times|W|$ binary matrix indicating the membership between segments in $S$ and waves in $W$, and the $\boldsymbol{\vec{s}}_W$ is the resulting vector indicating the significance of waves.
	
	Similarly, we can implement the decoding for the rating of wave-to-wave comparison by assigning the weight $\boldsymbol{\mathcal {S}}_{ij}$ to the corresponding wave pair $(w_p,w_q)$ if and only if $s_i \in w_p$ and $s_j\in w_q$ hold. Let us denote the rating of wave pairs as a matrix $\boldsymbol{\mathcal{W}}$, and write the membership matrix $\boldsymbol{\mathcal{M}}$ as a set of its column vectors (i.e., $\{\boldsymbol{\mathcal{M}}\}=\{\vec{\mathbf{m}_p}\}$ where the $i^{th}$ element  of $\vec{\mathbf{m}_p}$ indicates if $s_i\in w_p$). We have
	\begin{equation}
		\boldsymbol{\mathcal{W}}_{pq}=\vec{\mathbf{1}}^\top\left(\underbrace{\Big[\vec{\mathbf{m}_p}(\vec{\mathbf{m}_q})^\top\Big]}_{s_i\in w_p\land s_j\in w_q}\odot\boldsymbol{\mathcal {S}}\right)\vec{\mathbf{1}},
	\end{equation}
	where $\boldsymbol{\mathcal{W}}_{pq}$ implies the significance of comparing two waves $w_p$ and $w_q$ to the inference.
	
	\subsubsection{Human Comprehensible Representations}
	With the ratings of the waves and pairs, we are able to inspect what kind of evidence has been referred for the decision making in the CNN model. While this improves the transparency and interpretability of the deep learning, the saliency maps are evidence generated more from a machine learning perceptive, which is not necessarily friendly enough to medical technicians.
	In this section, we convert the saliency maps into human comprehensible representations of decision trees/forests
	\footnote{The decision trees/forests are selected because they have been considered (among our medical technicians) as the best understandable representations of the inference logic for the classification tasks.}. 
	
	\begin{table*}
		\centering \caption{Performance comparison for base model selection. The best results are in bold font.} \label{tb:select_base_model}
		\begin{tabular}{l|cccccc|cccc}
			\hline
			Model&\makecell[c]{Accuracy \\ (\%)}&\makecell[c]{F1-score \\ (\%)}&\makecell[c]{Recall \\ (\%)}&\makecell[c]{Precision \\ (\%)} &\makecell[c]{TPR \\ (\%)} &\makecell[c]{TNR \\ (\%)} &\makecell[c]{Model \\ Complexity} &\makecell[c]{Efficiency \\ (s)} &\makecell[c]{Accuracy/ \\ Complexity} &\makecell[c]{Accuracy/ \\ Efficiency}\\ \hline
			ResNet18 	&71.85 	&71.22 	&71.09 	&71.52 	&77.71 	&64.47 	&\textbf{15.5}	&\textbf{1.93} &\textbf{0.05}	&\textbf{0.37}\\
			ResNet34 	&71.51 	&70.93 	&70.82 	&71.15 	&76.88 	&64.75 	&29	    &2.55	&0.02	&0.28\\
			ResNet50 	&71.61 	&70.73 	&70.55 	&71.43 	&\textbf{79.79} 	&61.32 	&64.2	&3.39	&0.01	&0.21\\
			VGG16 	    &72.29 	&71.63 	&71.49 	&71.99 	&78.50 	&64.47 	&146	&2.14	&0.01	&0.34\\
			DenseNet121	&\textbf{72.56} 	&\textbf{72.03} 	&\textbf{71.92} 	&\textbf{72.22} 	&77.51 	&\textbf{66.33} 	&22.7	&5.70	&0.03	&0.13\\
			\hline
		\end{tabular}
	\end{table*}
	
	{\textbf{Attributes:}} To this regard, first we need to establish the candidate attributes for tree/forest generation. Before we start, hereafter we 
	will use the terms ``wave'' and ``wave genre'' differently, because the $6$ waves used in cardiology (e.g., Fig.~\ref{fig:ECGwave}) are indeed wave genres while from a data point of view, the waves (e.g., those in the previous section) are instances of wave genres. Our candidate attributes are thus established on wave genres to be cardiac meaningful.
	Therefore, for the wave-level evidence, the candidates are straightforwardly all wave genres, while for the comparative evidence, the candidates are wave genre pairs.
	Furthermore, for a signal sample, which is usually composed with several wave instances for a genre, we need to summarize the instances for the genre using statistics (e.g., mean, standard deviation)\footnote{Among a wide range of features available for summarization, the mean and standard are selected, because it is a common agreement among our medical technicians that they are the most straightforward features.}.
	
	{\textbf{Construction:}} To construct the trees/forests, we can use the saliency maps as a guide for attribute selection. More specifically, we will select an appropriate subset of attributes from the candidate attributes, which can significantly reduce the size of the decision trees/forests into human comprehensive levels (i.e., with much less number of attributes and shorter heights)\footnote{The reduction has to be done here not only for representation generation. Computationally, selecting the optimal subset of attributes is a NP-hard combinational problem which is impractical to solve by brute-force search.}. 
	More specifically, the saliency information in the maps, which indicates which segments/pairs play more important roles than others, can be considered as a ranking information for selecting attributes from the candidates in an orderly manner. Therefore, we can control the scale of the evidence representations by controlling the maximum number of attributes $t_{max}$ together with the maximum height $h_{max}$ of the trees/forests.
	Furthermore, the inference logic will be formed in a friendly way for medical technicians through the tree branching.
	Denote the candidate attribute set as $\{a_i\}$ where $a_i$ can be either a wave genre or a pair of wave genres and all the $a_i$s are ranked in descending order according to their saliency values in the maps. This process can be formulated as
	\begin{gather}
		\{a_t\}^*|t_{max},h_{max}=\mathop{\boldsymbol{\displaystyle \argmax}}_{\{a_t\}\subseteq \{a_i\}} \hspace{0.1pt}\rho\Big(Tr(\{a_t\},h)\Big),\notag\\
		1 \leq t\leq t_{max},1 \leq h\leq h_{max}
	\end{gather}
	where the $\{a_t\}$ is a set consisting the first $t$ attributes in $\{a_i\}$, $\{a_t\}^*$ denotes the optimal set of attributes under the conditions of $t_{max}$ and $h_{max}$, $Tr(\cdot)$ is a tree/forest builder which takes a subset of attributes $\{a_i\}$ and the height of the tree/forest $h$ as the inputs, and the $\rho(\cdot)$ is a performance evaluator (e.g., Accuracy or F1-score for classification tasks).
	More details about the implementation can be found in Section~\ref{sec:experi_compre_evidence}, in which we will conduct experiments to generate and evaluate the trees/forests. More importantly, we can see that the resulting evidence representations can provide more and finer-scale cardiac evidence comparing with that in the previous medical studies.
	Some of these findings, which have not been reported in the previous studies of eating behavior, not only validate the effectiveness of the proposed method, but also serve as new evidence for cardiology analysis.

		\begin{figure*}[t]
		
		\centering
		\includegraphics[width=0.97\textwidth]{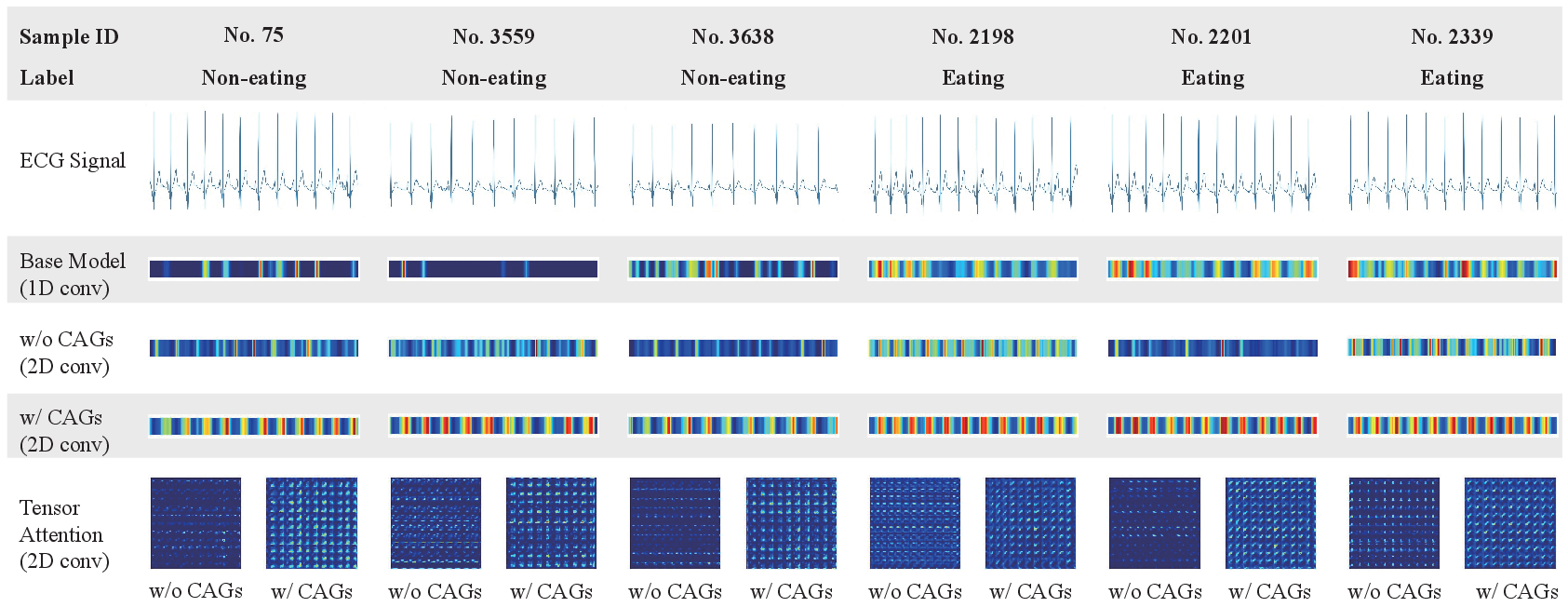}
		\caption{Examples of saliency maps generated by the Base Model (using the ECG signals as 1D inputs directly) and collocative learning (using the 2D-based collocative tensors as inputs). Maps by the collocative learning are generated at wave- and tensor-levels. Results of the collocative models with and without the periodic CAG regulations have been demonstrated respectively. It is easy to see that the collocative with CAGs has captured the periodic nature of the signals at either wave or tensor level. }\label{fig:ECGSamples}
	\end{figure*}

	\section{Experiments}
	\subsection{Setup}
	\subsubsection{Dataset}
	To further study the proposed method, we have employed a desensitized dataset collected by the West China Hospital of Sichuan University. It includes $150,037$ samples where each of them is an ECG segment with a duration of 10 seconds and has been labeled by medical technicians with either ``eating/digesting'' ($85,508$ samples) or ``non-eating'' ($64,529$ samples).
	%
	%
	The ``eating/digesting'' samples are selected and labeled by technicians for their regular meals according to the design of the corresponding medical study. Each label is a result based on the agreement of more than 2 (out of 3) technicians.
	The experiments in this section have been conducted on this Regular Meal dataset \footnote{The whole dataset in fact also comes with  $29,557$ samples for snacks and has been collected from the same of group of volunteers but not fully labeled, because the existing studies on eating are all based on regular meals. Furthermore, the labeled subset is with a less restrict setting (i.e., each sample has only been labeled by 1 technician). However, we will release the snack dataset for reference.}. 
	The desensitization has been done through anonymization and shuffling.
	To the best of our knowledge, this is the first and largest ECG dataset for eating behavior analysis with high quality labeling. Both the dataset and source code can be found at \emph{https://github.com/XXX (will open to public upon acceptance)}.
	
	\subsubsection{Evaluation}
	We use 10-fold cross validation where in each round, the dataset will be randomly divided into a training and a testing set consisting of $90\%$ and $10\%$ samples of the whole dataset, respectively.
	The standard metrics of accuracy, F1-score, Recall, and Precision have been employed for evaluating the performance of the classification (e.g., eating or non-eating). True Positive Rate (TPR), and True Negative Rate (TNR) have also been included for reference.
	
	\subsection{Selection of Base Model}
	The base model selection is not critical in the proposed framework in the sense that it is open and compatible with any CNN-based models. However, we still need a base model to conduct the large amount of experiments in the following sections for studying the characteristics of the proposed methods.
	To this end, we have fed the samples into a set of popular CNN models directly (without using any collocative learning processes) to study how these models fit to the problem. The models include ResNet18 \cite{he2016deep}, ResNet34 \cite{he2016deep}, ResNet50 \cite{he2016deep}, VGG16 \cite{simonyan2014very}, and DenseNet121 \cite{huang2017densely}.
	In addition to the standard metrics, we have included Model Complexity (the number of parameters), and Efficiency (time used in seconds for predicting the labels of samples in the testing set), together with two results of Accuracy/Complexity and Accuracy/Efficiency for evaluating how a model is balanced between the performance and cost.
	
	The results are shown in Table~\ref{tb:select_base_model}. The DenseNet121 has obtained the best performance over others. However, the performance gaps among models are in fact minor indicated by the mean F1-scores at $71.31$ with a narrow standard deviation of $0.53$.
	Surprisingly, a better performance is not guaranteed by increasing the complexity of the model. This is particularly obvious if we compare the three ResNet models. The best performance has been obtained by using the simplest model of ResNet18. This holds true in the comparison between the VGG16 and DenseNet121 as well. We consider this is an indication that the CNN models do not fit for the purpose directly.
	Finally, we can see the ResNet18 is the best balanced model which gives a satisfactory performance (with F1-score approaching the mean) and best efficiency ($9.81\%\sim 66.14\%$ faster than others). Therefore, we will select ResNet18 as the ``Base Model'' to continue the study in the rest sections of this paper.

	\begin{table}
		\centering \caption{Performance comparison for collocative tensor construction with different metrics. The best results are in bold font.} \label{tb:single_and_multi_view}
		\setlength\tabcolsep{3pt}
		\begin{tabular}{l|cccccc}
			\hline
			Model& 	\makecell[c]{Accuracy\\(\%)}&	\makecell[c]{F1-score\\(\%)}&	\makecell[c]{Recall\\(\%)}&	\makecell[c]{Precision\\(\%)} &	\makecell[c]{TPR\\(\%)} &	\makecell[c]{TNR\\(\%)}\\ \hline
			Base Model &71.85 	&71.22 	&71.09 	&71.52 	&77.71 	&64.47 \\
			Euclidean &72.74 	&72.45 	&\textbf{72.51} 	&72.41 	&74.53 	&70.49 \\
			Manhattan &72.38 	&72.11 	&72.19 	&72.05 	&73.77 	&\textbf{70.62} \\
			Cosine	&70.82 	&70.01 	&69.86 	&70.55 	&78.33 	&61.38 \\
			Mahalanobis	&70.38 	&69.85 	&69.78 	&69.98 	&75.10 	&64.45 \\
			Max	&70.84 	&69.75 	&69.58 	&70.79 	&80.60 	&58.57 \\
			Avg	&72.84 	&72.15 	&71.99 	&72.59 	&79.41 	&64.57 \\
			Min	&72.69 	&71.91 	&71.72 	&72.52 	&80.17 	&63.28 \\
			Multi-View	&\textbf{73.57}	&\textbf{72.56}	&72.32	&\textbf{73.76}	&\textbf{83.26}	&61.39 \\
			\hline
		\end{tabular}
	\end{table}

        \begin{table}
		\centering \caption{Performance comparison between the collocative models with and without periodic CAG regulation. The best results are in bold font.} \label{tb:cag_table}
		\begin{tabular}{l|cccccc}
			\hline
			Model& 	\makecell[c]{Accuracy\\(\%)}&	\makecell[c]{F1-score\\(\%)}&	\makecell[c]{Recall\\(\%)}&	\makecell[c]{Precision\\(\%)} &	\makecell[c]{TPR\\(\%)} &	\makecell[c]{TNR\\(\%)} \\ \hline
			{Base Model} &71.85 	&71.22 	&71.09 	&71.52 	&77.71 	&64.47\\ 
			\hline
			w/o CAGs&73.57	&72.56	&72.32	&73.76	&\textbf{83.26}	&61.39\\ 
			{Impr. (\%)} & 2.39 & 1.88 &1.73 &3.13 &7.14&-4.78 \\\hline
			w/ CAGs 	&\textbf{74.38}	&\textbf{73.72}	&\textbf{73.53}	&\textbf{74.21}	&80.92	&\textbf{66.13}\\
			Impr. (\%) & 3.52 & 3.51 &3.43 &3.76 &4.13 &2.57 \\ 
			\hline
		\end{tabular}
	\end{table}
	
\begin{table*}
		\centering \caption{Performance comparison with the state-of-the-art.} \label{tb:sota}
		\setlength\tabcolsep{10pt}
		\begin{tabular}{c|l|cccccc}
			\hline
			Group &Model& 	\makecell[c]{Accuracy\\(\%)}&	\makecell[c]{F1-score\\(\%)}&	\makecell[c]{Recall\\(\%)}&	\makecell[c]{Precision\\(\%)} &	\makecell[c]{TPR\\(\%)} &	\makecell[c]{TNR\\(\%)} \\ \hline
			\multirowcell{2}{Traditional Machine Learning} 
			& Wave\_SVM 	&68.98 	&66.99 	&67.05 	&69.62 	&\textbf{83.96} 	&50.14 \\ 
			&Comp\_SVM 	&71.01 	&69.69 	&69.53 	&71.24 	&82.49 	&56.57 \\ \hline
			\multirowcell{5}{Sophisticated CNNs}  &ResNet18 	&71.85 	&71.22 	&71.09 	&71.52 	&77.71 	&64.47 \\ 
			&ResNet34 	&71.51 	&70.93 	&70.82 	&71.15 	&76.88 	&64.75 \\ 
			&ResNet50 	&71.61 	&70.73 	&70.55 	&71.43 	&79.79 	&61.32 \\
			&VGG16 	&72.29 	&71.63 	&71.49 	&71.99 	&78.50 	&64.47 \\
			&DenseNet121 	&72.56 	&72.03 	&71.92 	&72.22 	&77.51 	&66.33 \\ \hline
			Recurrent Models   
			&LSTM 	&72.42 	&71.73 	&71.57 	&72.16 	&79.02 	&64.12 \\ \hline
			\multirowcell{3}{Ad-hoc Deep Models}
			&iResNet	&72.75 	&71.77 	&71.56 	&72.79 	&81.98 	&61.10 \\ 
			&Spect\_CNN 	&73.45 	&73.12 	&73.15 	&73.10 	&75.80 	&\textbf{70.49} \\ 
			&CNN+LSTM	&72.13 	&71.19 	&70.99 	&72.06 	&80.91 	&61.07 \\ \hline
			\multirowcell{2}{Ours}  
			&w/o CAGs	&73.57	&72.56	&72.32	&73.76	&83.26	&61.39 \\
			&w/ CAGs	&\textbf{74.38} 	&\textbf{73.72} &\textbf{73.53} &\textbf{74.21} &80.92 &66.13 \\
			\hline
		\end{tabular}
	\end{table*}
	
	\subsection{Collocative Tensor Construction}
	\label{sec:experi_tensor_construct}
	In this section, we are interested in searching the best configuration for the collocative tensor construction. More specifically, we have constructed the tensors by calculating the collocative relation of two ECG segments with standard metrics of Euclidean distance, Manhattan distance, Cosine similarity, and Mahalanobis distance respectively. In addition, we have tried representing the relation by using the Maximum, Average, and Minimum values of two segments respectively. Finally, we generate a multi-view tensor by concatenating these single-view tensors following Eq.~(\ref{eq:concat_single_view}). The results are shown in Table~\ref{tb:single_and_multi_view}.

	\subsubsection{Comparative Relation Modeling}
	We can see that by transforming the problem from unary relation modeling into binary (comparative) relation modeling using collocative learning, the performance has been improved from the Base Model by using most of the relation metrics, and 4 metrics (out of 8), namely the Euclidean, Manhattan, Avg, and Multi-view), have even outperformed the DenseNet121 (which has obtained the best performance in the base model selection).
	However, we only consider the performance gain as a ``by-product'' of introducing collocative learning. As we have discussed in Section~\ref{sec:evidence_backtracking}, the key advantage of the proposed methods is to bring possibility for evidence backtracking on the comparative features, so that human understandable knowledge can be extracted for more effective pathological studies. This will be investigated specifically in the experiments in Section~\ref{sec:experi_compre_evidence}.
	
	\subsubsection{Multi-View vs. Single-View}
	Comparing between multi- and single-view tensor constructions, multi-view has obtained the best performance on all metrics except the TNR, resulting from the complementary information encapsulated in its member (single-) views. This can be found by looking at the performance distributions of the single- and multi-view runs.
	More specifically, single-view tensors are usually performing well on some metrics while showing inferiority on others. For example, the Max has achieved the best performance (among single-view tensors) on TPR but work defectively on other metrics.
	Another example is the performance of Avg which is approaching the best among single-view runs on all metrics except TNR. 
	By contrast, the superiority of multi-view tensor over singe-view tensors is consistent over almost all metrics. Its effectiveness has been obtained by integrating the single-views for a more comprehensive representation.

	\subsection{Periodic Attention Regulation}
	In this section, we further evaluate the effectiveness of CAGs for period regulation. The results are shown in Table~\ref{tb:cag_table}. It is not surprising that by adding the CAGs as the periodic regulator, the performance of the proposed method has been further improved by $3.52\%$ ($3.51\%$) in Accuracy (F1-score) when compared to that of the Base Model (the multi-view learner).
	In addition, its superiority over the Base Model has been validated on all metrics, including the TNR on which the multi-view learner has a slightly degraded performance. Let us zoom in for more details.

	
	\begin{figure*}[t]
		
		\centering
		\includegraphics[width=1.\textwidth]{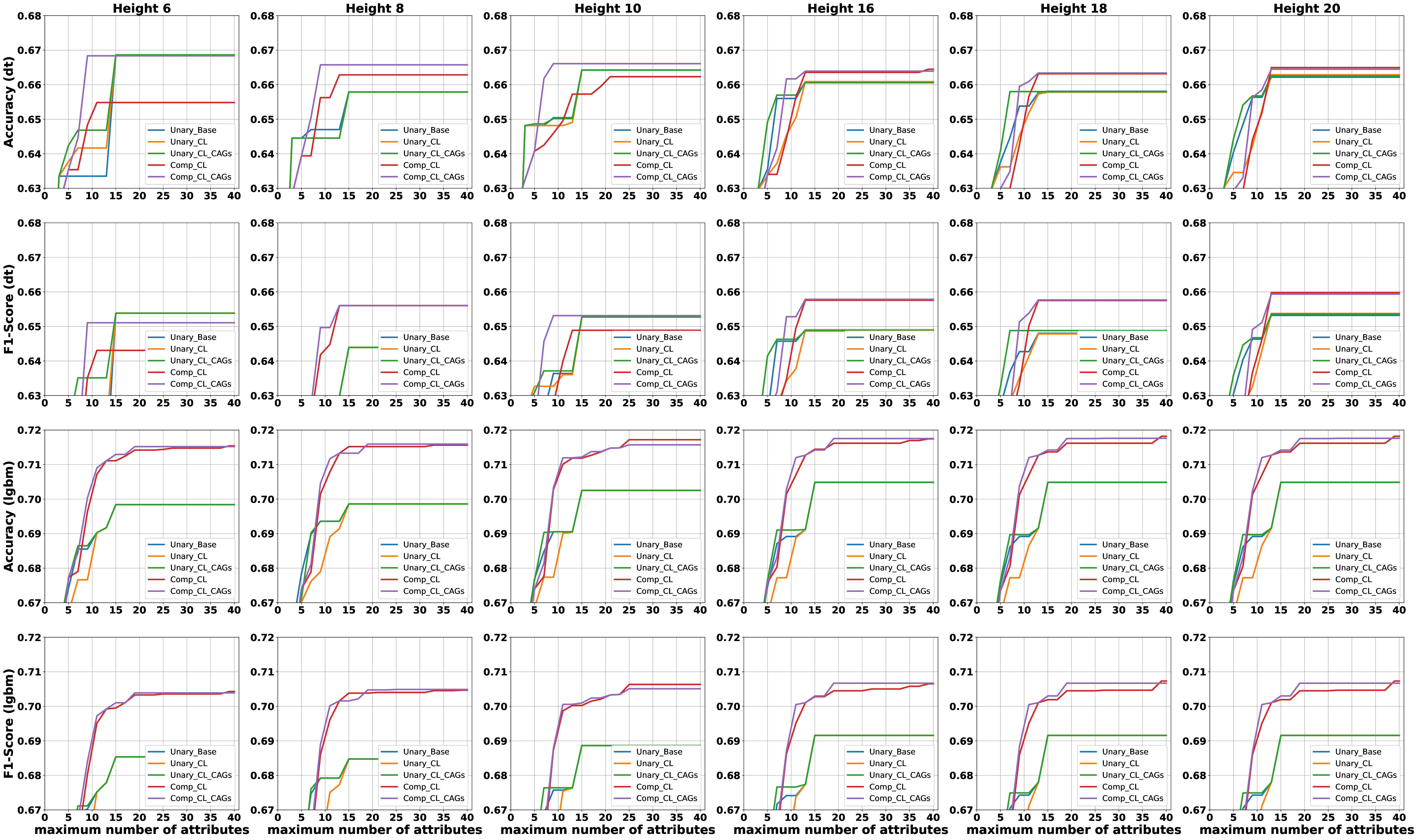}
		\caption{Performance evaluation of decision trees (dt) and LightGBM forests (lgbm) generated with different configurations. The decision trees/forests generated with the comparative attributes learned though the collocative learning (i.e., the Comp\_CL\_CAGs) have outperformed the others significantly.}\label{fig:cag_models_fig}
	\end{figure*}
    
	\subsubsection{Period Modeling}
	In Fig.~\ref{fig:ECGSamples}, we have shown several typical examples of the ECG signals and the corresponding saliency maps generated by the Base Model (1D convolutional model) and the Collocative Learning (2D convolutional models with and without CAGs), respectively.
	It is clear to see that the collocative learning with CAGs has captured the periodic characteristics of the original signals with superior performance over the Base Model and the one without CAGs.

	At wave-level, the Base Model focuses on much fewer segments of non-eating signals than those of the eating signals. This is good for building discriminative inference models, which explains why the Base Model can still obtain an acceptable performance (with F1-score of $71.22\%$) in the classification task. However, the selection of segments does not appear to have a fixed pattern in terms of signal periods. It will limit its application in evidence backtracking for medical studies. We will show how this affect the human comprehensible evidence generation later in Section~\ref{sec:experi_compre_evidence}. 
	By contrast, the attention patterns of the collocative learning (either with or without CAGs) has been well established, in the sense that collocative learning without CAGs puts its focus more on the ST intervals of the signals, and the one with CAGs focuses more on the PR intervals for non-eating signals and the ST intervals for eating signals.
	More importantly, we can find that the attention of the collocative learning with CAGs has formed a periodic pattern much more obvious than that of the one without CAGs.
	
	This becomes more intuitive when one looks at the 2D tensor saliency maps of the two collocative learning models. The periodic and diagonal patterns of the model with CAGs are distributed more evenly and with more uniformed shapes in the maps than those without CAGs, which has verified the effectiveness of the periodic and diagonal regulator of CAGs. Furthermore, the 2D saliency maps have provided much more comparative features than the 1D signals. We will see this is useful for generating human comprehensible evidence in Section~\ref{sec:experi_compre_evidence}. 
	
	\subsection{Performance Comparison with the State-of-the-art}
	To verify if the proposed method is a state-of-the-art (SOTA), we compare it with $4$ groups of $11$ existing methods as follows.
	\begin{itemize}
		\item Traditional machine learning: Wave\_SVM \cite{cortes1995support} and Comp\_SVM \cite{cortes1995support} are selected because they are indeed more popular (than deep models) in ECG analysis and with recognized performance.
		\item Sophisticated CNN for images: ResNet18 \cite{he2016deep}, ResNet34 \cite{he2016deep}, ResNet50 \cite{he2016deep}, VGG16 \cite{simonyan2014very}, and DenseNet121 \cite{huang2017densely} are selected for their popularity.
		\item Recurrent models: LSTM \cite{hochreiter1997long} is selected to represent the deep models for sequential data and it has also been widely adopted for ECG processing \cite{hou2019lstm,saadatnejad2019lstm,yildirim2018novel}.
		\item Ad-hoc deep models: iResNet \cite{yang2020multi} is a representative for 1D convolutional networks which is a modification of ResNet for processing the ECG data specifically; Spect\_CNN \cite{ullah2020classification} has employed 2D CNN for ECG processing in frequency domain (on spectrum); CNN+LSTM \cite{chen2020automated} is a deep model for ECG processing which has been designed to take the advantage of CNN for representation learning and that of LSTM for sequential data processing.
		
	\end{itemize}
	The results are shown in Table~\ref{tb:sota}. Not surprisingly, our collocative learning framework (w/ CAGs) has outperformed all $11$ SOTA methods with performance gain $1.27\%\sim 7.83\%$ ($0.82\%\sim 10.05\%$) in Accuracy (F1-score). Therefore, its performance is validated.
	For future work, active learning \cite{wei2012coaching,wei2011coached} can be integrated into the framework to optimize the annotation process for ECG data, iteratively boosting the capability of the system. 
    Additionally, leveraging generative models \cite{zhang2024compositional,zhang2024generative} could address the challenge of data scarcity by synthesizing high-quality training data.

	\section{Generation of Human Comprehensible Evidence Representations}
	\label{sec:experi_compre_evidence}
	As mentioned, the key advantage of the periodic collocative learning is to provide a way to guide the deep learning to conduct the inference on the comparative features with periodic attention. We consider this not only an effort towards interpretable deep learning, but also a new way for human comprehensible evidence building.
	We verify this by generating the decision trees/forests using the method we have introduced in Section~\ref{sec:relation_model}.
	

    \begin{figure*}[t]
		
		\centering
		\includegraphics[width=0.9\textwidth]{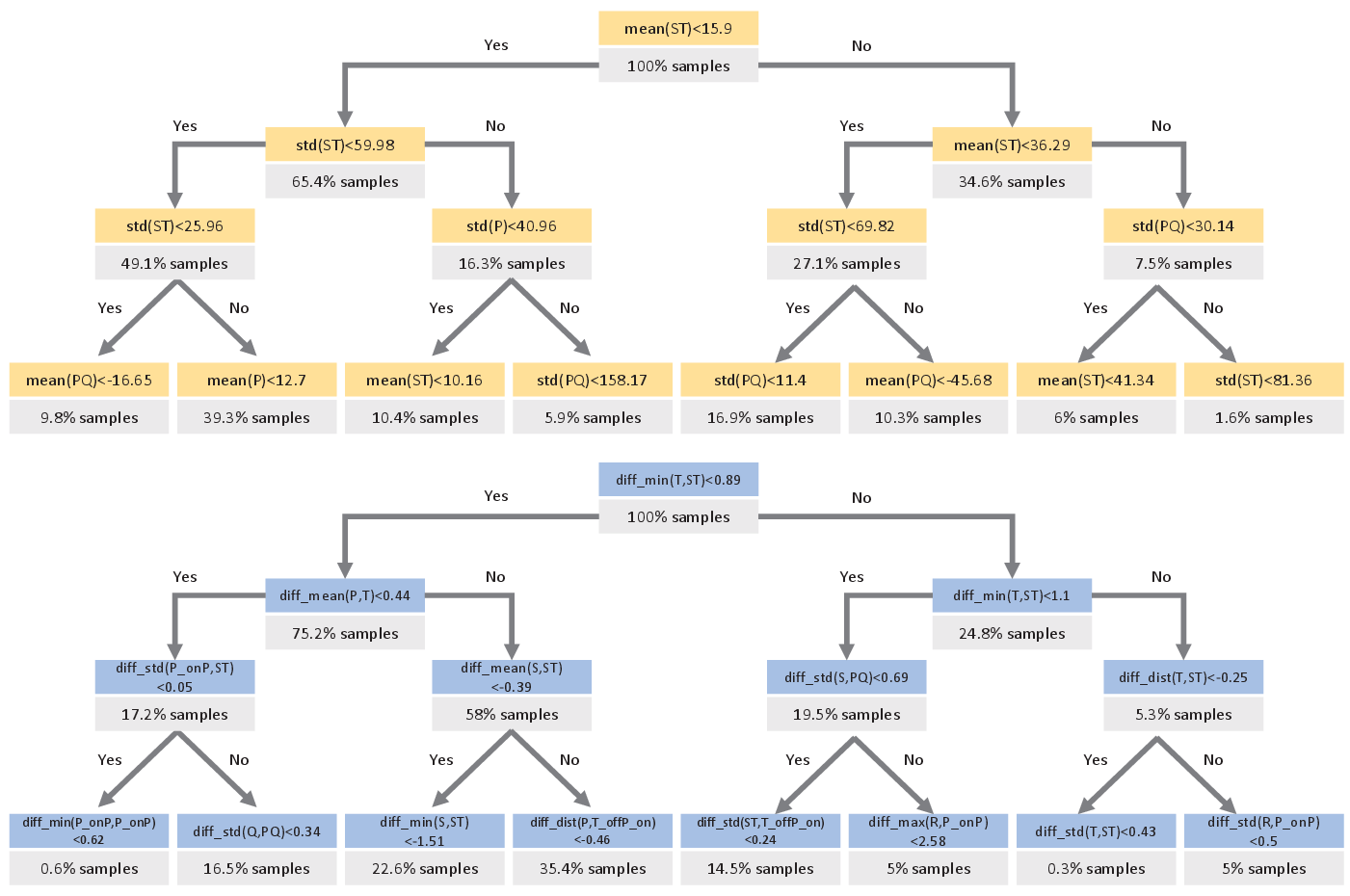}
		\caption{Examples of decision trees generated with the attributes learned by the collocative learning with periodic CAG regulation. The tree with yellow nodes has been generated with the unary attributes (i.e., Unary\_CL\_CAGs) and the one with blue nodes has been generated with the comparative attributes (i.e., Comp\_CL\_CAGs).}\label{fig:trees}
	\end{figure*}
    
	\subsection{Experimental Setting}
	We will further use the saliency maps to guide the construction of decision trees/forests. 
	From the machine learning perspective, we have implemented the decision trees in the most conventional way (using entropy as the splitting function). This is the simplest model but with the best comprehensible outputs (i.e., the exact knowledge in the form of trees). However, it is too simple to model complicated relations among attributes. In this case, a set of attributes will outperform the other only if it is with significant discriminability over the other. 
	Therefore, we have also implemented the decision forests with LightGBM \cite{ke2017lightgbm}, which can model much more complex relations, and thus do not rely on the discriminability of individual attributes that much. By comparing the performance of attributes on the decision trees and forests, we can study the discriminability of the attributes and the relations among them at the same time.

	To establish the candidate attributes, we divide the $6$ cardiac wave genres into finer scale for a more precise analysis, which results in $15$ subwave genres as the attributes representing unary features of the signals. Note that for the ST subwaves, we are indeed extracting the section from the S valley to T peak, because this is the most possible way to detect it automatically. Similarly, the PQ subwaves are extracted from the section from the P peak to Q valley. Consequently, we have $120$ genre pairs as the comparative candidates.
	Then we can use the (tensor- or wave-level) saliency maps as ranking information on these two groups of attributes respectively to generate the decision trees/forests. It results in $5$ configurations for the representation generation as shown in Table~\ref{tb:cag_description}.
	We will compare the performance of the decision trees/forests generated with these configurations with the assumption that the better discriminability a tree/forest obtains, the better capacity of generating a valid and comprehensible representation the corresponding model has. We will also discuss some typical cardiac evidence found through these representations by comparing them to previous medical research findings.
	
	\begin{table}
		\centering \caption{Configurations for decision tree/forest generation.} \label{tb:cag_description}
		\setlength\tabcolsep{4pt}
		\begin{tabular}{l|c|c|c}
			\hline
			Configuration  &Model& \makecell[c]{Attribute \\ Type}& Saliency Map \\ \hline
			Unary\_Base & Base Model & Unary & Wave-level\\ \hline
			Unary\_CL &w/o CAGs & Unary& Wave-level\\ \hline
			Unary\_CL\_CAGs	&w/ CAGs & Unary& Wave-level\\ \hline
			Comp\_CL &  w/o CAGs & Comparative & Tensor-level\\ \hline
			Comp\_CL\_CAGs& w/ CAGs & Comparative & Tensor-level\\ 
			\hline
		\end{tabular}
	\end{table}

	\subsection{Collocative vs. Conventional Learning}
	With the $5$ configurations, we feed them to the decision tree/forest learners (i.e., the entropy-based tree leaner and the LightGBM) with two parameters controlling the scale of the resulting evidence representations, namely the maximum height of trees $h_{max}\in[1,20]$ and the maximum number of attributes $t_{max}\in[1,40]$. This results in $800$ runs in total.
	Some of the results are shown in Fig.~\ref{fig:cag_models_fig}. We can find that the comparative attributes generated by collocative learning (either Unary\_CL\_CAGs or Comp\_CL\_CAGs) has significantly outperformed those by the conventional learning (Unary\_Base). The superiority of Comp\_CL\_CAGs over others is dominating and consistent wherever it is obtained by the tree or forest learner, and also valid for different combinations of parameters. 
	The superiority at wave-level is less obvious than that at tensor-level, because either there are only limited number of candidate attributes (i.e., $15$) for the configurations to make enough difference from each other, or the difference exists only in a narrow range from $5$ to $12$ (along the x-axis) and is less visible when one observes the results at the full range of $0$ to $40$. 
	However, the performance of comparative configurations by collocative learning (Comp\_CL and Comp\_CL\_CAGs) have demonstrated notable dominance over the unary configuration by conventional learning (Unary\_Base) either with the decision trees or forests (LightGBM) which has confirmed the effectiveness of collocative learning in terms of representation generation.
	
	\begin{table}
		\centering \caption{The ranking lists of attributes (pairs) generated by the collocative learning with periodic CAG regulations (i.e., Comp\_CL\_CAGs). Only top-40 comparative attributes are shown due to space limitation.} \label{tb:ranking}
		\setlength\tabcolsep{2pt}
		\begin{tabular}{l|l}
			\hline
			\makecell[l]{Attribute Type}  & \makecell[c]{Ranking Lists  of Attributes (Pairs) \\in Descending Order} \\ \hline
			Unary & \makecell[l]{TP, ST, P\_onP, PQ, P, R, S,T, Q, TT\_off, head, \\Tail, QR, RS, other} \\ \hline
			Comparative & \makecell[l]{(ST,TP), (TP,TP), (ST,ST),(P\_onP,TP), (PQ,TP), \\(P\_onP,ST), (PQ,ST),(R,TP), (P,TP), (S,TP), 
				\\(Q,TP), (T,TP), (head,TP),(TT\_off,TP), (R,ST), \\(S,ST), (P,ST),(Q,ST), (T,ST), (ST,TT\_off),
				\\(ST,head),(P\_onP,PQ), (Tail,TP), (R,P\_onP), \\(S,P\_onP),(Q,P\_onP), (P,P\_onP), (T,P\_onP),
				\\(P\_onP,TT\_off), (R,PQ), (P,PQ), (S,PQ),(Q,PQ), \\(P\_onP,P\_onP), (T,PQ), (ST,Tail),
				(PQ,TT\_off), \\(QR,TP), (P\_onP,head), (PQ,head)}
			\\ \hline
		\end{tabular}
	\end{table}

	\subsection{With vs. Without CAG Regulation}
	Comparing between the collocative learning with and without the CAG regulation, the superiority of the CAG regulation is obvious and consistent on the unary configurations (i.e., Unary\_CL\_CAGs over Unary\_CL) either with simple decision trees or complex forests.
	For the comparative configurations, the performance is different on the decision trees and forests.
	On the trees, the advantage of 
	Comp\_CL\_CAGs over Comp\_CL is more obvious on the short trees where the models are much simpler (e.g., at the heights of $6$, $8$, and $10$ in Fig.~\ref{fig:cag_models_fig}). While on the higher trees (e.g., at the heights of $16$, $18$, and $20$), it is still obvious when the maximum number of attributes is in the range of $[1,13]$ but later the performance of the two models coverage into the same level. 
	It is an indication that with the CAG regulation, the learner can find the effective attributes more efficiently than without it. However, since the collocative attributes itself is effective enough for representation generation, the configuration without the assistance of the CAG regulation (i.e., Comp\_CL) can also obtain the acceptable performance in the more complex models (i.e., when the trees become higher and the maximum number of attributes is larger) in which the relations among attributes can play a more important role so as to compensate the lack of discriminability of the individual attributes. The convergence of the performance indicates both configurations have reached the upper bound of representation generation with decision trees.
	For the same reason, the superiority of the CAG regulation on the forest models (LightGBM) is more consistent over the two parameters, because the forests are much more complex learners than the trees so that both models are at the same level of leveraging the relations among attributes, and thus the advantages of collocative learning with CAGs can be brought into full swing by generating better discriminability of the attributes.

	\subsection{Typical Representations and Cardiac Evidence}
	In Table~\ref{tb:ranking}, we have shown the lists of both unary and comparative attributes ranked by our collocative learning with CAGs.
	The ST and TP sections, which are highly ranked, are also frequently refereed by the decision trees/forests (see Fig.~\ref{fig:trees}). This is in fact consistent with the findings in previous medical research, which concluded that the eating/digesting mainly affects the QT section of the ECG signals, arising from postprandial increase in cardiac output and the effect of c-peptide and glucose on cardiac repolarization.
	Moreover, the previous medical findings refer to the QT as the section from the start point of Q waves and the end of T wave. This is a coarse-scale reference as it has covered 4 out 6 cardiac waves of a period.
	By contrast, we can see from Table~\ref{tb:ranking} and Fig.~\ref{fig:trees} that the proposed method has located the evidence in a much finer scale. More importantly, we are able to provide readable rules in the form of decision trees/forests.
	From the examples of evidence representations in Fig.~\ref{fig:trees}, it is much intuitive for medical technicians/researchers to analyze the interactions between the ECG and the eating behavior than using only coarse-scale statistics.
	

	\section{Conclusion}
	In this paper, we have presented a pilot study of AEM using 24-hour ECG. A collocative learning framework has been proposed for this problem, which has addressed the issue of adapting the sophisticated 2D image-based deep models to 1D ECG signal analysis. More importantly, the framework is capable of modeling the human logic of cardiac analysis based on comparative features, and therefore, the decision trees/forests can be generated as human comprehensive representations of the inference evidence. The effectiveness of the framework has been validated on the largest ECG dataset for eating behavior, where the superior performance over SOTA has been observed. The capability of cardiac evidence mining has also been verified through the consistency of the evidence backtracked by the framework and that of the previous medical studies.

    \bibliographystyle{IEEEtran}
    {\small
    \bibliography{tmi-2022-reference.bib}
    }
\end{document}